\documentclass[11pt]{article}

\usepackage[final]{acl}

\usepackage{times}
\usepackage{latexsym}
\usepackage{float} 
\usepackage{array}
\usepackage{tabularx}
\usepackage{dblfloatfix}
\usepackage[T1]{fontenc}
\usepackage[utf8]{inputenc}

\usepackage{microtype}

\usepackage{inconsolata}

\usepackage{graphicx}
\usepackage{amsfonts}
\usepackage[most]{tcolorbox}
\definecolor{mygreen}{RGB}{0,120,0} 
\tcbset{before skip=4pt, after skip=4pt}
\usepackage{enumitem}
\usepackage{amsmath}
\usepackage{booktabs}  
\usepackage{multirow}

\title{\textsc{CoCoGEC}: Counterfactual Generation for Robust Grammatical Error Correction}

\author{
 \textbf{Qianyu Wang},
 \textbf{Xiaoman Wang},
 \textbf{Yuanyuan Liang},
 \textbf{Xinyuan Li},
 \textbf{Yunshi Lan\thanks{Corresponding author}}
\\
 East China Normal University
\\
 \{wangqianyu, xmwang, leonyuany, xyli\}@stu.ecnu.edu.cn, yslan@dase.ecnu.edu.cn
}

\begin{document}
\maketitle

\begin{abstract}
Grammatical error correction (GEC) systems are usually trained and evaluated on GEC benchmarks, but their performance often drops sharply once the surrounding context is slightly perturbed or extended.
This indicates that the existing GEC models usually fail to understand the error patterns in the varying contexts.
In this paper, we thoroughly investigate the counterfactuals for GEC tasks, where the subtle changes to the contexts could lead to the label flipping issue.
We propose \textsc{CoCoGEC}, a counterfactual generation framework that creates copies of training instances with error-irrelevant contexts altered.
Our framework systematically generates counterfactuals by (1) generating intra- and inter-sentence counterfactuals that maintain the error patterns as well as syntax of the original instances by altering the word-level and sentence-level contexts; (2) revising the generated counterfactuals by selecting the instances with flipped labels and high GEC Mutual Information (MI) coefficient.
Extensive experiments show that our method substantially improves the stability of GEC models, outperforming a set of data augmentation baselines.
Particularly, it could achieve absolute $F_{0.5}$ gains of $+9.9$, $+11.3$, and $+20.8$ points on the perturbed BEA-19*,CoNLL-14*, and TEM-8* data set.Our code is released at \url{https://github.com/Quinnok/CoCoGEC}.
\end{abstract}

\section{Introduction}
\label{sec:intro}



Grammatical Error Correction (GEC) aims to automatically detect and correct grammatical errors in text, supporting applications such as intelligent writing assistants and computer-assisted language learning.
It has attracted increasing attention from both academia and industry in recent years \citep{katinskaia-yangarber-2023-grammatical,katinskaia-yangarber-2024-gpt,li-etal-2025-explanation,kovalchuk2025omnigec}.
However, we observe a substantial gap between the well-trained GEC models and their applications to the real world.
\begin{figure}[ht]
    \centering
    \includegraphics[width=\linewidth]{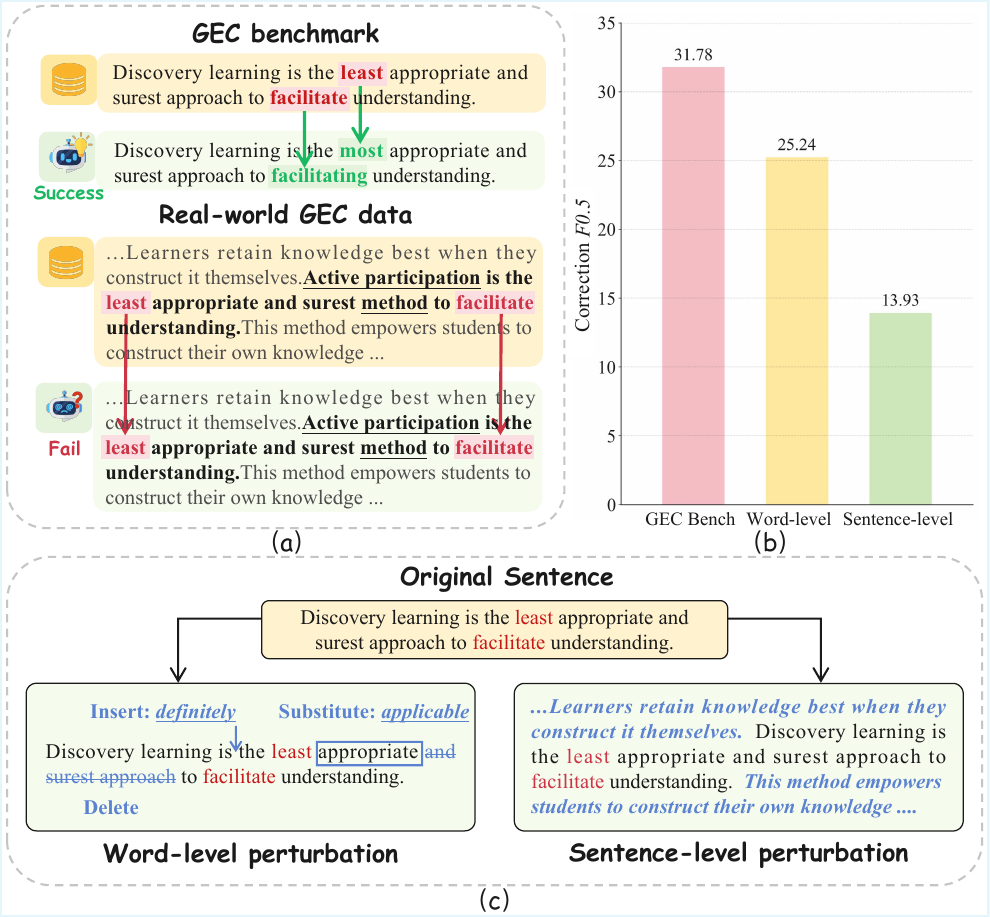}
    \caption{Motivation for \textsc{CoCoGEC}. (a) Context shift between standard benchmarks and real-world inputs. (b) Robustness drop on TEM-8 under perturbations with GPT-4. (c) The illustrative examples of two types of counterfactuals. }
    \label{fig:intra_inter_perturb}
\end{figure}

Figure~\ref{fig:intra_inter_perturb} (a) illustrates an example from the BEA-19 task where ``\textit{least}''
should be corrected to ``\textit{most}'' and ``\textit{facilitate}'' to
``\textit{facilitating}'' in the context of ``\textit{Discovery
learning}''. While GPT-4 correctly revises this sentence on standard GEC benchmarks, GEC models often fail when encountering similar errors in diverse real-world contexts. For example, in a longer passage about ``\textit{Active participation}'', the same erroneous phrase reappears. However, GPT-4
leaves it unchanged and thus exhibits under-correction errors.


To quantify this robustness gap, we conduct a preliminary experiment. We generate the "real-world" data via two augmentation methods: (1) word-level perturbation, involving random token alterations in test sentences from TEM-8~\citep{tem8-standard}; and (2) sentence-level perturbation, constructed by randomly combining sentences from the test set\footnote{Implementation details of the preliminary experiment can be found in Appendix~\ref{app:prompts}.}. 
As shown in Figure~\ref{fig:intra_inter_perturb} (b), there is an observable performance drop between the original GEC data and "real-world" data, especially for sentence-level perturbation. 

Prior studies ~\citep{zhang-etal-2023-robustgec,wang-etal-2024-improving-grammatical}
have revealed that current GEC models are vulnerable to seemingly harmless perturbations.
But these studies mainly focus on noise-based attacks or broad augmentation, rather than interpreting the potential perturbations that fundamentally affect GEC models.
For example, \citeauthor{wan-etal-2020-improving}~(\citeyear{wan-etal-2020-improving}) and \citeauthor{park-etal-2023-synthetic-alone}~(\citeyear{park-etal-2023-synthetic-alone}) use noise injection,
\citeauthor{lichtarge-etal-2019-corpora}~(\citeyear{lichtarge-etal-2019-corpora}) and \citeauthor{stahlberg-kumar-2021-synthetic}~(\citeyear{stahlberg-kumar-2021-synthetic}) generate pseudo corpora,
and \citeauthor{wang-etal-2024-improving-grammatical}~(\citeyear{wang-etal-2024-improving-grammatical}) and \citeauthor{li-lan-2025-large}~(\citeyear{li-lan-2025-large}) propose contextual augmentation.
These approaches expand or re-distribute training data, but the perturbations are often random or coarsely controlled, limiting their ability to explain the performance gap.

In this paper, we address robustness to word- and sentence-level perturbations by asking: \textit{how can we make a GEC model focus more on error patterns while ignoring varying context?}
To this end, we propose a novel \textbf{CoCoGEC} method, inspired by counterfactual analysis.
The intuition behind \textsc{CoCoGEC} is to create copies of training instances with their error-irrelevant contexts altered.
We identify two types of decoupled counterfactuals for GEC data, which aim to alter the word-level and sentence-level contexts without influencing the original error patterns in a sentence, but could confuse the prediction of a GEC model.
We display the "counterfactual" GEC in Figure~\ref{fig:intra_inter_perturb} (c).
With the counterfactuals, the GEC model would learn to put more emphasis on the error patterns when learning how to correct a sentence.

\textsc{CoCoGEC} uses large language models (LLMs) to generate span-controlled intra-sentence variants that substitute error-irrelevant spans while keeping the gold correction edits valid. It also constructs inter-sentence variants by attaching coherent, error-free prefixes and suffixes to emulate discourse-level context shifts. We enforce an edit-level fidelity constraint to filter invalid candidates, then rank the remaining counterfactuals with a GEC mutual-information score and keep the most challenging ones for augmentation.
The experimental results consistently verify that \textsc{CoCoGEC} improves the robustness of GEC models.

The main contributions of this work are:
\begin{itemize}
  \item To the best of our knowledge, this is the first study to explore counterfactuals for GEC tasks by characterizing potential perturbations with three criteria.
  \item We introduce \textsc{CoCoGEC}, a counterfactual generation pipeline tailored to contexts in GEC, which systematically constructs intra-sentence and inter-sentence variants without influencing the original error pattern, but could confuse the prediction of a GEC model.
  \item We propose a novel GEC mutual information coefficient that captures the dependence between the varying context and the model predictions for identifying high-quality counterfactuals.
  \item We demonstrate on the RobustGEC benchmark that COCOGEC consistently improves robustness under both intra- and inter-level perturbations, without sacrificing performance on standard test settings.
\end{itemize}
\section{Related Work}
\label{sec:related}

\paragraph{Robust GEC.}
Modern GEC systems mainly fall into sequence-to-sequence generation~\citep{vaswani2017attention,junczys2018lowresource}, sequence-to-edit correction~\citep{awasthi2019pie,stahlberg2020seq2edits,omelianchuk2020gector,qorib-ng-2023-system} (including hybrid detection--correction variants~\citep{li-etal-2023-templategec,li-wang-2024-detection}), and recent LLM-based pipelines with prompting or light supervision~\citep{loem-etal-2023-exploring,coyne-etal-2023-analyzing,katinskaia-yangarber-2024-gpt,tang-etal-2024-ungrammatical}. 
Despite strong benchmark performance, existing models can be brittle under small contextual shifts, motivating robustness-oriented training and data construction.
Robustness is typically pursued along two complementary directions. Model-centric methods improve stability by explicitly regularizing invariance—via adversarial objectives~\citep{dang-xie-liu-2021-adversarial}, distillation-style constraints~\citep{xia-etal-2022-kd-cgec}, or consistency-based post-training on constructed variants and hard cases (e.g., RobustGEC/TemplateGEC/CLEME2.0/CSA)~\citep{zhang-etal-2023-robustgec,li-etal-2023-templategec,ye-etal-2024-cleme2,tang-etal-2023-csa}. 
Data-centric methods instead broaden supervision by synthesizing training pairs through noise injection~\citep{solyman-etal-2023-jksuci,sun-etal-2023-syntactic-cognitive}, back-translation~\citep{fang-etal-2023-transgec}, and contextual or edit-based augmentation~\citep{wang-etal-2024-improving-grammatical,ye-etal-2023-mixedit}, sometimes coupled with robustness-oriented annotation or curricula~\citep{li-lan-2025-large,zhang-etal-2025-loss-aware}. 
However, much of this augmentation primarily targets error diversity or reweighting, and indiscriminate synthetic data may even degrade GEC performance~\citep{park-etal-2023-synthetic-alone}. 
 In contrast, our approach is data-centric: we generate context-decoupled counterfactuals with an edit-subset constraint ($E' \subseteq E$) to target context robustness, and they can be used with various GEC backbones.


\paragraph{Counterfactual Analysis Beyond GEC.}
Counterfactual data augmentation (CDA) improves robustness by generating controlled perturbations that preserve or systematically modify labels, encouraging models to rely on invariant features and generalize out of distribution~\citep{wang-etal-2024-survey,jiang-etal-2024-robust-ce}. 
Recent progress largely comes from strengthening controllability and label fidelity in generation: diffusion-based frameworks provide a powerful mechanism for robust synthesis and transfer~\citep{xin2024diffusion,chen2024diffusioncls,bae2025salad,wang2023cra}, while optimization-driven formulations enforce invariance through reinforcement learning, information bottlenecks, and contrastive objectives~\citep{chen2021reinforced,sreedhar2025metric,chang2024ceib,choi2022c2l}. 
In parallel, counterfactuals have shifted from rule-based edits to more controllable generative pipelines, including distillation and LLM-driven synthesis~\citep{chen-etal-2023-disco,nguyen2024llms,howard-etal-2022-neurocounterfactuals,treviso2023crest,zhou2023implicit}, as well as explanation-oriented designs that improve interpretability and faithfulness~\citep{yang2023relationbased,an2025selfexplaining}. 

Collectively, these studies offer broad tools for building context-invariant NLP models, but they focus on classification tasks, leaving counterfactual generation for structured prediction and GEC comparatively underexplored. Our work brings CDA to GEC by designing contextual counterfactuals tailored to error correction, rather than label-flipping counterfactuals commonly used in discriminative settings.

\section{Method}
\label{sec:method}
\label{sec:controllable-cf}
\begin{figure*}[t]
    \centering
    \includegraphics[width=1\linewidth]{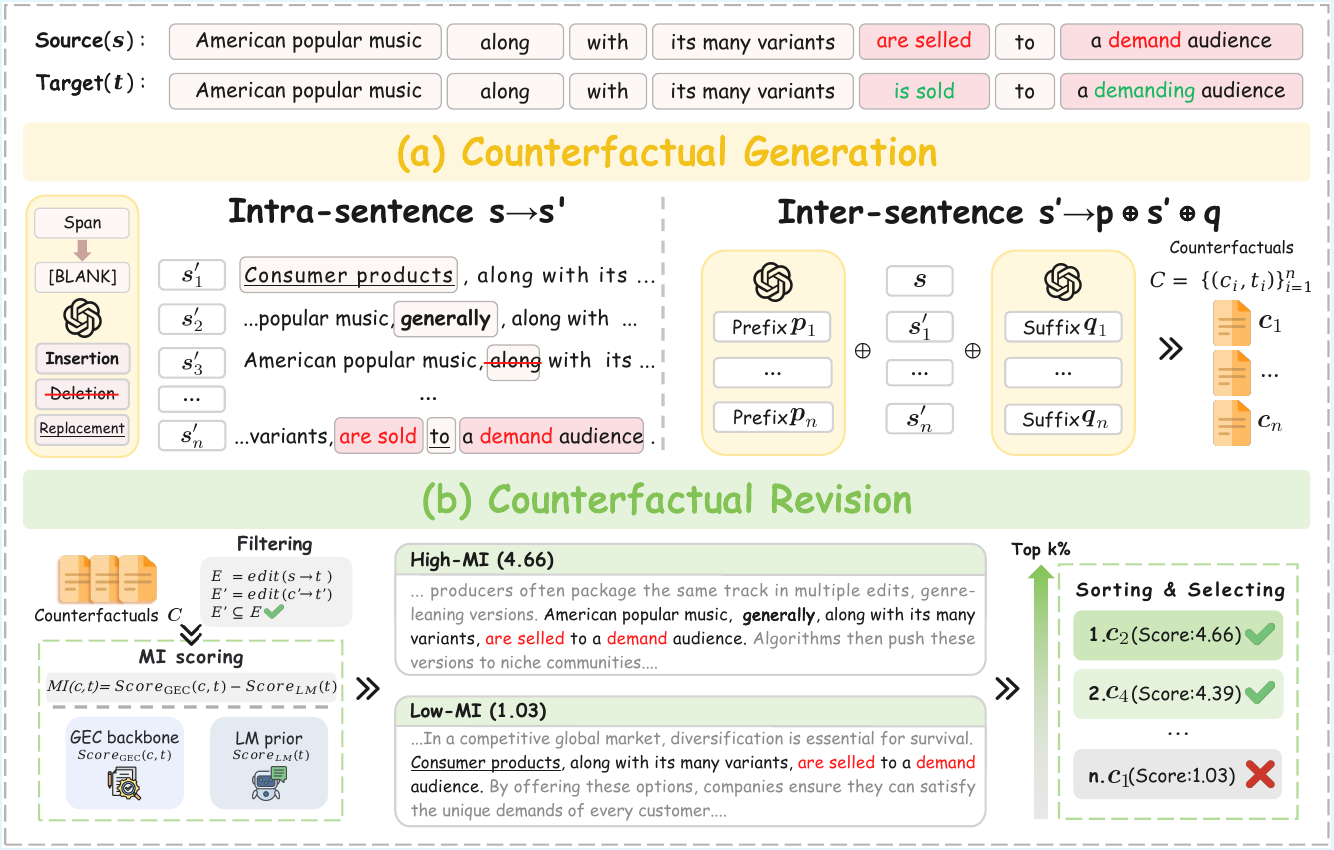}
    \caption{Overview of \textsc{CoCoGEC}: we generate span-controlled intra-sentence variants and attach coherent prefix/suffix to form long-context counterfactuals, then filter by edit-set consistency and rank by GEC mutual information to keep the most confusing yet valid augmentations.}
    \label{fig:placeholder}
\end{figure*}

\subsection{Definition of Counterfactuals for GEC}
\label{subsec:cf-definition}
Existing studies~\cite{wang:arxiv2024,verma:compsurvey2024} have made a formal definition of counterfactuals in general machine learning tasks.
A counterfactual example $c$ usually disturbs a model to predict an instance $x$ as an alternative class $y'$ instead of its original class $y$ by making \textit{minimal yet necessary} changes to $x$ as follows:
\begin{align*}
     &arg\operatorname*{min}_{c} \quad \text{dist}(x, c) \\
    \quad&\text{s.t.} \quad  f(c) \neq f(x)
\end{align*}

where $f$ is a task-specific model $f: \mathtt{X} \in \mathbb{R}^d \rightarrow \mathtt{Y}$ to bridge the mapping from $x$ to $y$ and $dist{(\cdot, \cdot)}$ is a distance function that measures the cost of changes required to alter the prediction.

The above definition outlines the fundamental principles of the counterfactual generation problem.
Considering the distinct problem formulation of the GEC task, we identify the counterfactuals for GEC as conducting a subtle change to the source text, but resulting in a subset of the original edits.
Motivated by the example in Figure~\ref{fig:intra_inter_perturb}, we mainly focus on the counterfactuals for intra-sentence and inter-sentence. 
We consider a counterfactual to be shown in the form $c = p \oplus s' \oplus q$, where $s'$ minimally modifies the $s$, and $p$ and $q$ are a prefix and a suffix to the source text, respectively.
As a result, we have:
{
\begin{align}
    & \operatorname*{argmin}_{c} \quad \text{dist}(s, c) \nonumber \\
    \quad & = \text{syntax\_dist}(s, s') + \text{semantic\_dist}(s', p \oplus q) \nonumber \\
    &\text{s.t.} \quad \mathcal{E}' \subseteq \mathcal{E} \label{eq:counter_gec}
\end{align}
}

Formally, $(s, t)$ denotes the annotated source and target text for grammatical error correction.
Ideally, $f(\cdot)$ is a GEC model which takes the source text as the input and perfectly produces the corrected text, such that $f(s) =t$ and $f(c) = t'$.
The edit mapping of the original text $s \rightarrow t$ and counterfactual text $c \rightarrow t'$ are denoted as $\mathcal{E}$ and $\mathcal{E}'$, respectively.

Regarding Equation~(\ref{eq:counter_gec}), we interpret the countertfactuals for GEC as follows:
\begin{itemize}[leftmargin=*, itemsep=1pt]
    \item \textbf{Minimal syntactic revision to source text}. For the revision in intra-sentence, we make minor verbal adjustments, which may change the semantics of the sentence rather than its syntax.
    We denote it as minimal $syntax\_dist(s, s')$.
    \item \textbf{Semantic coherence to the revision}. For the revision in inter-sentence, we append a prefix and a suffix to the revised source text, but keep semantic coherence to the revised source text, which prevents the illogical flow of the counterfactuals.
    We define minimal $semantic\_dist(s', p \oplus q)$.
    \item \textbf{Flipped edit labels}. We consider that a good counterfactual for GEC could flip the prediction of a GEC model, where the original edits cannot be recognized well.
    To avoid intertwined grammatical errors, new edits cannot be introduced.
    Hence, we deem $\mathcal{E}'$ as a subset of $\mathcal{E}$ as the outcomes of the counterfactuals.
\end{itemize}

\subsection{Counterfactual Generation with LLMs}
\label{subsec:cf-generation}

Next, we generate counterfactuals following the above definition. 
We first target at constructing the syntactically similar $s'$ to $s$, which could flip the original error annotations.
However, it is not trivial to control the perturbation of the source sentence.
Unlike the label-flipping objective in discriminative settings, our goal in GEC is to \textit{preserve} the original grammatical errors and \textit{not include} new errors.
Thus, we would like the error-irrelevant context to be perturbed while keeping the syntax unchanged.

We propose a pipeline of counterfactuals generation for GEC with the integration of intra-sentence and inter-sentence perturbation, which corresponds to $s'$ and $p \oplus q$ defined in Equation~(\ref{eq:counter_gec}), accordingly.
Specifically, we first distill counterfactuals for $s'$ generation with LLMs in a controllable manner.
Then, we generate $p$ and $q$ for the purpose of the inter-sentence perturbation.
At last, we extract the counterfactuals satisfying our objective of flipping the original edits while not introducing new edits.
This pipeline results in an integrated counterfactual in the format of $p \oplus s' \oplus q$.



\paragraph{Intra-sentence counterfactual generation of $s'$.}
To generate $s'$, we target at minimal syntactic revision to source text $s$.
Prior study~\cite{chen-etal-2023-disco} introduces $\mathcal{DISCO}$ method to prompt LLMs via in-context learning to generate counterfactuals for natural language inference, which controls the spans for variant generation.
We follow the principle of $\mathcal{DISCO}$ and distill intra-sentence counterfactuals with LLMs as follows:
\begin{itemize}[leftmargin=*]
    \item We first segment each source sentence $s$ into a sequence of candidate spans $\mathcal{W} = \{w | w \in s\}$ using a Flair-based chunker~\citep{akbik-etal-2019-flair}.
    To prevent unexpected edits to the erroneous regions, we discard all spans overlapping with $\mathcal{E}$, keeping only spans outside the gold edits.
    \item For each retained span $w \in \mathcal{W}$, we sample a masked variant by replacing $w$ with a special token \texttt{[BLANK]}, and feed into an LLM with an instruction asking the model to fill the blank with an alternative phrase, which achieves insertion, deletion, or replacement for $w$.
    We generate multiple variants for each $s$.
\end{itemize}

As shown in Figure~\ref{fig:placeholder}, the illustrative example ``\textit{American popular music, along with its many variants, are selled to a demand audience. }'' can have multiple intra-sentence counterfactuals with the unchanged syntax. 
To preserve alignment between the perturbed source and its correction, we apply LLM filling only to non-error spans. Whenever a span $w$ in the source sentence $s$ is replaced by an LLM-generated fragment $\tilde{w}$, we simultaneously replace the aligned span in the gold target $t$ with the same $\tilde{w}$. As a result, we collect a set of counterfactual sentence pairs, denoted as $(s', t')$.




\paragraph{Inter-sentence counterfactual generation of $p \oplus q$.}
To probe long-range context and expose GEC models to paragraph-level correction, we generate prefix and suffix for $s'$.
To ensure the semantic coherence of the revision, for each pair $(s', t')$, we automatically attach error-free context on both sides by sampling a short grammatical prefix $p$ and suffix $q$ from an LLM.
As the example shown in Figure~\ref{fig:placeholder}, the intra-sentence counterfactual is mentioned in the middle of long contexts.
Eventually, we obtain $c=p \oplus s' \oplus q$ as defined in Equation~(\ref{eq:counter_gec}).
We attach the same prefix and suffix to $t'$, and the sentence pair $(c, t')$ forms a pre-defined counterfactual.
It is worth noting that since $p$ and $q$ are error-free contexts, the edit mapping $c \rightarrow t'$ will not introduce new errors beyond the span $s$.
We denote the set of generated counterfactuals as $\mathcal{C} = \{(c_i, t_i')\}_{i=1}^{n}$


\subsection{Counterfactual Revision with GEC Mutual Information}
\label{sec:mi-ranking}
\paragraph{Edit-subset counterfactual filtering.}
To ensure that label flipping occurs in $c$ within the span $s$, we conduct filtering.
Specifically, we employ ERRANT~\citep{bryant-etal-2017-automatic} to conduct the edit mapping of $s \rightarrow t$ and $c \rightarrow t'$ to produce $\mathcal{E}$ and $\mathcal{E}'$, respectively.
If a sentence pair satisfies $\mathcal{E}' \subseteq \mathcal{E}$, we keep it; otherwise, we abandon it.This results in a set of counterfactuals that follows our pre-defined principle.

\paragraph{Mutual-information scoring and selection.}
The above steps ensure a counterfactual to be \textit{valid} but not \textit{optimal}.
An optimal counterfactual in GEC tasks should be a hard negative that is highly related to the correct form but particularly challenging for a GEC model to distinguish.
In other words, the flipped edits should be difficult to detect via a GEC model.
In Figure~\ref{fig:placeholder}(b), we show two counterfactual candidates derived from the source sentence about ``\textit{American popular music}''.
The retained candidate makes only a minor contextual insertion (e.g., ``\textit{generally}'') while keeping the original subject, forming a challenging near-miss that remains compatible with the gold target.
By contrast, another candidate changes the subject to ``\textit{Consumer products}'', which becomes semantically misaligned with the original correction target and is therefore ranked low and filtered out.
To turn the large pool of generated counterfactuals into high-quality examples.
We propose a novel Mutual-Information–based scoring function to measure the GEC mutual information of a counterfactual.
Previous studies on counterfactual data augmentation~\citep{chen2018learning,plyler-chi-2025-icda} adapted Mutual Information (MI) to counterfactual generation, holding the following assumption:

\begin{itemize}[leftmargin=*]
\item Seeking a counterfactual is referred to as the maximum mutual information criterion:
\begin{align*}
  I(c; y)= \mathbb{E}_{\mathtt{C}, \mathtt{Y}}[\log \frac{P_{\mathtt{C,Y}}(c, y)}{P_{\mathtt{C}}(c)P_{\mathtt{Y}}(y)} ]
\end{align*}

where $c$ denotes the counterfactual and $y$ denotes the original prediction. $I(c; y)$ measures the dependence between the counterfactual and the original prediction.
\end{itemize}

When it comes to the GEC task, we tend to measure the dependence between the erroneous source text $c$ and the original target text $t$ as follows:
\begin{align*}
    \text{argmax}_{c \in \mathcal{C}} I(c; t) &= \mathbb{E}_{\mathtt{C}, \mathtt{T}}[\log \frac{P_{\mathtt{C,T}}(c, t)}{P_{\mathtt{C}}(c)P_{\mathtt{T}}(t)} ] \\
     & = \mathbb{E}_{\mathtt{C}, \mathtt{T}}[\log P_{\mathtt{T|C}}(t | c) - \log P_{\mathtt{T}}(t)]
\end{align*}


As we can see from the formula, a good counterfactual $c$ in the GEC task should have a high probability $P_{\mathtt{T|C}}(t | c)$ of transferring a counterfactual $c$ to the original target text $t$, where the perturbed errors in $c$ do not alter the original prediction effectively.
This is also influenced by the fluency of the prediction $t$, a smaller $\log P_{\mathtt{T}}(t)$ indicates a less fluent $t$, which makes the GEC model more confused.

In GEC tasks, we approximate these two terms using neural network-based GEC models.
We employ a GEC model with a Seq2seq framework and compute the joint probability of the sequential tokens in the target text:
\begin{align*}
    \log P_{\mathtt{T|C}}(t | c) = \frac{1}{|t|} \sum_{i=1}^{|t|}
\log \text{GEC}_{\text{Seq2seq}}\bigl(w_i \mid c, w_{<i}\bigr),
\end{align*}
where $w_i$ denotes the generation of $i$-th token in $t$.

For the GEC model with the Seq2Edit framework, we approximate with the joint probability of the sequential operations leading to the target text:
\begin{align*}
    \log P_{\mathtt{T|C}}(t | c) = \frac{1}{|c|} \sum_{i=1}^{|c|}
\log \text{GEC}_{\text{Seq2edit}}\bigl(e_i \mid c, e_{<i}\bigr)
\end{align*}
where $e_i$ denotes the operation to $i$-th token in $c$.

Regarding $\log P_{\mathtt{T}}(t)$, we employ GPT-2-medium~\citep{radford-2019-gpt2}  to compute the joint probability of the tokens in the target text.
\begin{align*}
    \log P_{\mathtt{T}}(t) = \frac{1}{|t|} \sum_{i=1}^{|t|}
\log \text{LM}\bigl(w_i \mid w_{<i}\bigr)
\end{align*}
where $w_i$ denotes the generation of $i$-th token in $t$.

We compute an MI score for each counterfactual in $\mathcal{C}$, sort them in descending order, and select the top $k$ percent to form the final set. 
MI scoring reflects our desiderata: a high-quality counterfactual is a close near-miss to the correct form rather than a random, malformed utterance.

\begin{table*}[t]
\small
\centering
\resizebox{\textwidth}{!}{
\begin{tabular}{lc|ccc|ccc|ccc}
\toprule
& & \multicolumn{3}{c|}{BEA-19*} & \multicolumn{3}{c|}{CoNLL-14*} & \multicolumn{3}{c}{TEM-8*} \\
Method & Data Size & Prec. & Rec. & $F_{0.5}$    & Prec. & Rec. & $F_{0.5}$    & Prec. & Rec. & $F_{0.5}$  \\
\midrule
GECToR-large & - 
& $26.86$ & $24.54$ & $26.36$
& $22.28$ & $21.31$ & $22.08$
& $39.51$ & $34.27$ & $38.35$ \\
+ CPR method & 36K 
& $28.32$ & $27.55$ & $28.16$
& $18.09$ & $20.51$ & $18.54$
& $49.81$ & $22.88$ & $40.32$ \\
+ $\mathcal{DISCO}$ & 42K
& $31.49$ & $21.05$ & $28.65$
& $26.88$ & $17.79$ & $24.38$
& $47.27$ & $26.46$ & $40.83$ \\
+ TypeDA  & 30K
& $26.87$ & $25.64$ & $26.63$
& $23.87$ & $22.30$ & $23.54$
& $39.51$ & $36.89$ & $38.96$ \\
+ \textsc{CoCoGEC}  & 25K
& $\mathbf{51.55}$ & $11.27$ & $\mathbf{30.07}$
& $\mathbf{34.52}$ & $12.22$ & $\mathbf{25.29}$
& $\mathbf{52.90}$ & $31.76$ & $\mathbf{46.77}$ \\

\midrule
T5-large & - 
& $27.27$ & $38.22$ & $28.92$
& $23.64$ & $25.80$ & $24.04$
& $23.18$ & $38.39$ & $25.18$ \\
+ CPR method  & 36K 
& $27.29$ & $38.70$ & $28.99$
& $25.91$ & $21.63$ & $24.93$
& $24.47$ & $38.76$ & $26.41$ \\
+ $\mathcal{DISCO}$ & 42K
& $27.70$ & $39.28$ & $29.43$
& $24.06$ & $26.76$ & $24.56$
& $24.85$ & $39.33$ & $26.83$ \\
+ TypeDA  & 30K
& $27.96$ & $41.42$ & $29.90$
& $24.16$ & $27.72$ & $24.80$
& $25.64$ & $41.19$ & $27.74$ \\
+ \textsc{CoCoGEC}  & 25K
& $\mathbf{32.93}$ & $31.52$ & $\mathbf{32.64}$
& $\mathbf{27.61}$ & $22.43$ & $\mathbf{26.40}$
& $\mathbf{33.16}$ & $35.02$ & $\mathbf{33.51}$ \\
\midrule
Qwen3-8B & - 
& $26.68$ & $37.34$ & $28.29$
& $22.22$ & $23.72$ & $22.50$
& $31.11$ & $46.44$ & $33.32$ \\
+ CPR method  & 36K 
& $27.48$  & $40.64$  & $29.38$ 
& $25.77$  & $18.43$  & $23.88$ 
& $33.25$  & $44.94$  & $35.07$  \\
+ $\mathcal{DISCO}$ & 42K
& $26.20$ & $38.01$ & $27.93$
& $21.56$ & $27.40$ & $22.52$
& $36.64$ & $46.98$ & $38.33$ \\
+ TypeDA  & 30K
& $25.94$ & $48.40$ & $28.59$
& $21.05$ & $31.41$ & $22.54$
& $27.02$ & $51.69$ & $29.88$ \\
+ \textsc{CoCoGEC}  & 25K
& $\mathbf{36.63}$ & $46.12$ & $\mathbf{38.22}$
& $\mathbf{36.78}$ & $25.64$ & $\mathbf{33.83}$
& $\mathbf{55.70}$ & $48.50$ & $\mathbf{54.09}$ \\
\midrule
Qwen3-4B & - 
& $22.14$ & $22.11$ & $22.13$
& $13.08$ & $9.29$  & $12.10$
& $24.07$ & $24.34$ & $24.13$ \\
Qwen3-14B & - 
& $26.86$ & $39.57$ & $28.70$
& $23.27$ & $27.40$ & $23.99$
& $33.48$ & $54.49$ & $36.28$ \\
Qwen3-235B & - 
& $37.01$ & $43.82$ & $38.20$
& $29.88$ & $34.62$ & $30.72$
& $46.88$ & $64.79$ & $49.63$ \\
GPT-4o & - 
& $34.43$ & $48.89$ & $36.60$
& $33.72$ & $34.10$ & $33.80$
& $45.60$ & $65.97$ & $48.60$ \\
LLaMA3-8B  & - 
& $23.09$ & $44.52$ & $25.55$
& $19.3$ & $29.01$ & $20.68$
& $16.24$ & $48.30$ & $18.72$ \\
\bottomrule
\end{tabular}
}
\caption{Main results on RobustGEC test sets.
We denote the perturbed data subsets in RobustGEC as BEA-19*, CoNLL-14*, and TEM-8*. 
\textsc{CoCoGEC} consistently improves $F_{0.5}$ over CPR, $\mathcal{DISCO}$ generation, and TypeDA across all backbone--dataset pairs, with the largest absolute gains on the long-context TEM-8* data. }

\label{tab:main_results}
\end{table*}

\begin{table*}[t]
\small
\centering
\resizebox{\textwidth}{!}{%
\begin{tabular}{l|ccc|ccc|ccc|ccc}
\toprule
& \multicolumn{3}{c|}{Source}
& \multicolumn{3}{c|}{Word-level Perturbation}
& \multicolumn{3}{c|}{Sentence-level Perturbation}
& \multicolumn{3}{c}{Combined Perturbation} \\
Method
& Prec. & Rec. & $F_{0.5}$
& Prec. & Rec. & $F_{0.5}(\Delta \downarrow)$
& Prec. & Rec. & $F_{0.5}(\Delta \downarrow)$
& Prec. & Rec. & $F_{0.5}(\Delta \downarrow)$ \\
\midrule
GECToR
& $62.4$ & $38.7$ & $55.6$
& $58.6$ & $38.3$ & $53.0 \, (2.6)$
& $38.4$ & $33.6$ & $37.3 \, (18.3)$
& $39.5$ & $34.4$ & $38.3 \, (17.3)$ \\
+ CPR
& $60.4$ & $39.8$ & $54.7$
& $58.1$ & $40.3$ & $53.4 \, (1.3)$
& $39.8$ & $34.2$ & $38.5 \, (16.2)$
& $40.8$ & $29.9$ & $38.0 \, (16.7)$ \\
+ TypeDA
& $62.9$ & $40.9$ & $56.8$
& $60.1$ & $36.8$ & $53.3 \, (3.4)$
& $40.1$ & $33.6$ & $38.6 \, (18.2)$
& $47.3$ & $26.5$ & $40.9 \, (15.9)$ \\
+ $\mathcal{DISCO}$
& $69.1$ & $37.3$ & $59.0$
& $65.7$ & $37.3$ & $57.0 \, (2.0)$
& $48.2$ & $30.6$ & $43.3 \, (15.9)$
& $39.5$ & $36.9$ & $39.0 \, (20.1)$ \\
+ \textsc{CoCoGEC}
& $66.3$ & $45.2$ & $\mathbf{60.6}$
& $64.6$ & $46.5$ & $\mathbf{59.9 \, (0.7)}$
& $63.7$ & $32.5$ & $\mathbf{53.4 \, (7.2)}$
& $52.90$ & $31.76$ & $\mathbf{46.7 \, (14.0)}$ \\
\bottomrule
\end{tabular}
}
\caption{Performance breakdown on TEM-8 across disturbance settings. $\Delta\downarrow$ denotes the drop from the Source setting, computed as $F_{0.5}^{\text{Source}} - F_{0.5}^{\text{Perturbed}}$; lower $\Delta$ indicates better robustness.}
\label{tab:breakdown}
\end{table*}

\section{Experimental Setup}
\subsection{Dataset}
We conduct all experiments on RobustGEC \citep{zhang-etal-2023-robustgec}, which augments BEA-19, CoNLL-14, and TEM-8 with robustness-oriented perturbations to error-irrelevant context. Each original sentence pair has several human-generated variants, and we split the data by case into training, development, and test sets in a 7:1:2 ratio, keeping all variants of a case in the same split. For long-context evaluation, we use GPT-4\footnote{\url{https://openai.com/research/gpt-4}} to add a shared prefix $p$ and suffix $q$ to each source--target pair, yielding long-context test sets denoted BEA-19$^\ast$, CoNLL-14$^\ast$, and TEM-8$^\ast$. Unless otherwise noted, models are trained and tuned only on the original sentence pairs.

\begin{table*}[htbp]
\centering
\resizebox{\textwidth}{!}{%
\begin{tabular}{l|ccc|ccc|ccc|ccc|ccc}
\toprule
\multirow{2}{*}{\textbf{Model}} & \multicolumn{3}{c|}{ATK1} & \multicolumn{3}{c|}{ATK2} & \multicolumn{3}{c|}{ATK3} & \multicolumn{3}{c|}{ATK4} & \multicolumn{3}{c}{ATK5} \\
 & $F_{0.5}$ & TR$\uparrow$ & SR$\uparrow$ & $F_{0.5}$ & TR$\uparrow$ & SR$\uparrow$ & $F_{0.5}$ & TR$\uparrow$ & SR$\uparrow$ & $F_{0.5}$ & TR$\uparrow$ & SR$\uparrow$ & $F_{0.5}$ & TR$\uparrow$ & SR$\uparrow$ \\
\midrule
Qwen3-8B
& $7.98$ & $27.39$ & $27.39$
& $15.44$ & $38.59$ & $27.05$
& $18.73$ & $30.77$ & $16.26$
& $20.33$ & $23.54$ & $8.31$
& $21.29$ & $18.47$ & $3.90$ \\
Qwen3-8B +\textsc{CoCoGEC}
& $\mathbf{9.48}$ & $\mathbf{37.40}$ & $\mathbf{37.40}$
& $\mathbf{15.84}$ & $\mathbf{38.81}$ & $\mathbf{28.21}$
& $\mathbf{19.41}$ & $\mathbf{32.14}$ & $\mathbf{17.11}$
& $\mathbf{21.22}$ & $\mathbf{24.13}$ & $\mathbf{8.70}$
& $\mathbf{22.16}$ & $\mathbf{19.19}$ & $\mathbf{4.17}$ \\
\bottomrule
\end{tabular}%
}
\caption{Robustness to perturbation number from $1$ to $5$. \textsc{CoCoGEC} maintains a consistent advantage over the baseline across most attack settings, indicating improved robustness under stronger perturbations.}
\label{tab:erroratk}
\end{table*}

\begin{table}[t]
\centering
\resizebox{\columnwidth}{!}{%
\begin{tabular}{l|ccccc}
\toprule
Method & Prec. & Rec. & $F_{0.5}$ & CRS$\uparrow$ & P-CRS$\uparrow$ \\
\midrule
GECToR
& $39.51$ & $34.27$ & $38.35$ & $6.41$ & $64.30$ \\
\midrule
\multicolumn{6}{c}{\textit{Counterfactual Generation}} \\
\midrule
+ GPT-based $c$
& $40.31$ & $34.22$ & $39.00$ & $6.42$ & $65.36$ \\
+ $s'$
& $47.27$ & $26.46$ & $40.83$ & $10.45$ & $71.33$ \\
+ $p \oplus q$
& $45.79$ & $32.42$ & $42.33$ & $10.02$ & $66.50$ \\
\midrule
\multicolumn{6}{c}{\textit{Counterfactual Revision}} \\
\midrule
+ $k = 100\%$
& $44.05$ & $33.08$ & $41.31$ & $11.83$ & $67.93$ \\
+ $k = 30\%$
& $52.90$ & $31.76$ & $\mathbf{46.77}$ & $\mathbf{16.98}$ & $\mathbf{72.52}$ \\
\bottomrule
\end{tabular}}
\caption{Ablation of counterfactual components on TEM-8 with GECToR, showing that span-/document-level edits and MI-based selection ($k=30\%$) give the best $F_{0.5}$ and robustness.}
\label{tab:ablation}
\end{table}

\subsection{Evaluation Metrics}
We report standard edit-based metrics (precision, recall, and $F_{0.5}$) computed with ERRANT \citep{bryant-etal-2017-errant}. 
For RobustGEC, we additionally report CRS and P-CRS \citep{zhang-etal-2023-robustgec} to quantify correction consistency across context-perturbed variants (case-level and pairwise, respectively). 
For attacked sets, following CSA \citep{tang-etal-2023-csa}, we further report SR and TR, measuring recovery at the sentence and token levels; higher values indicate better robustness.

\subsection{Comparative Methods}
We compare \textsc{CoCoGEC} with representative robustness and augmentation baselines. 
As backbones, we use GECToR-large \citep{omelianchuk2020gector} for Seq2Edit, T5-large \citep{raffel2020t5} for Seq2Seq, and Qwen3-8B for LLM-based GEC. 
CPR \citep{zhang-etal-2023-robustgec} improves contextual consistency via KL-divergence regularization. 
$\mathcal{DISCO}$ \citep{chen2023disco} distills counterfactual data with LLMs.
TypeDA \citep{li-lan-2025-large} augments training data with type-aware LLM annotation through masked modeling and error filling. 
We also include zero-shot GEC baselines with GPT-4o \citep{achiam2023gpt4} and LLaMA3-8B as representative general-purpose LLMs.
\subsection{Implementation Details}

We generate counterfactuals by chunking candidate spans with Flair\footnote{\url{https://github.com/flairNLP/flair}} and enforcing edit constraints using gold edit sets extracted by ERRANT\footnote{\url{https://github.com/chrisjbryant/errant}}, followed by basic hygiene filtering (length control, empty/degenerate removal, and de-duplication). We estimate the likelihood term $\log P(t\mid c)$ with an ensemble of GEC scorers, and implement fine-tuning with LLaMA-Factory and LoRA~\citep{zheng2024llamafactory,hu2021lora}. Templates and remaining details are in Appendix~\ref{app:implementation}.

\section{Results and Analyses}
\label{sec:results}
\subsection{Main Results on RobustGEC}
\label{subsec:main_results}

Table~\ref{tab:main_results} reports the performance of \textsc{CoCoGEC} and data-augmentation baselines on the three RobustGEC benchmarks. We summarize three main observations:

\begin{itemize}[leftmargin=*]
    \item \textbf{\textsc{CoCoGEC} improves robustness across different backbones.}
    Across Seq2Edit (GECToR-large), Seq2Seq (T5-large), and LLM-based (Qwen3-8B) backbones, \textsc{CoCoGEC} consistently improves $F_{0.5}$ on all three benchmarks.
    For Qwen3-8B in particular, \textsc{CoCoGEC} yields absolute $F_{0.5}$ gains of $+9.9$, $+11.3$, and $+20.8$ points on BEA-19*, CoNLL-14*, and TEM-8*, respectively, indicating more context-invariant correction behavior, especially in long-context scenarios.

    \item \textbf{\textsc{CoCoGEC} outperforms existing augmentation methods.}
    Compared with noise injection (CPR), distillation-based augmentation ($\mathcal{DISCO}$), and type-aware augmentation (TypeDA), \textsc{CoCoGEC} attains the highest $F_{0.5}$ in every backbone--dataset pair in Table~\ref{tab:main_results}.
 On the context-perturbed TEM-8* benchmark in particular, it surpasses all baselines by a clear margin, suggesting that controllable context counterfactuals provide stronger training signals than random or loosely controlled perturbations, even when using fewer counterfactual instances.

    \item \textbf{\textsc{CoCoGEC} narrows the gap between compact models and large LLM baselines.}
Although large LLMs (e.g., GPT-4o, LLaMA3-8B, and Qwen3-235B) are strong zero-shot baselines, they can still be sensitive to contextual perturbations. When fine-tuned with \textsc{CoCoGEC}, Qwen3-8B matches or slightly surpasses these zero-shot LLM baselines on RobustGEC perturbed subsets, suggesting that counterfactual, data-centric optimization can be a parameter-efficient alternative to simply scaling model size.

\end{itemize}

\subsection{Breakdown by Perturbation Type}
\label{subsec:breakdown}
Since Section~\ref{subsec:main_results} shows consistent gains across GECToR, T5, and Qwen3, we use GECToR-large as a representative Seq2Edit backbone for a detailed robustness analysis on TEM-8. Table~\ref{tab:breakdown} further decomposes performance into the unperturbed \emph{Source} setting, word-level, sentence-level, and combined perturbations. Vanilla GECToR suffers the largest drop in $F_{0.5}$ under sentence-level and combined perturbations, while word-level perturbation alone is less harmful, suggesting that long-range contextual shifts are the main source of brittleness. Robustness-oriented augmentations reduce this drop, and \textsc{CoCoGEC} achieves the smallest $F_{0.5}$ degradation in all settings while maintaining strong source performance, indicating that training with context-decoupled counterfactuals stabilizes GEC in long-context scenarios.

\subsection{Ablation Studies}
\label{subsec:ablation}

We ablate \textsc{CoCoGEC} on TEM-8 with GECToR to disentangle the effects of counterfactual generation and revision.
As shown in Table~\ref{tab:ablation}, adding only global context rewriting (\texttt{+ GPT-based $c$}) yields small but consistent improvements, while intra-sentence variants $s'$ and inter-sentence expansion $p \oplus q$ each bring larger gains in $F_{0.5}$ and robustness metrics. The revision stage further improves performance: selecting a subset of counterfactuals with MI-based ranking ($k=30\%$) outperforms keeping all generated instances ($k=100\%$) in both $F_{0.5}$ and CRS/P-CRS\footnote{Definitions of CRS and P-CRS are given in Appendix~\ref{app:crs-metric}.}. Overall, this shows that selective filtering is crucial, as keeping all instances can dilute the gains. These results indicate that robustness gains come from controllable span edits combined with selective filtering, rather than simply enlarging the training set.

\begin{figure}[htbp]
    \centering
    \includegraphics[width=1\linewidth]{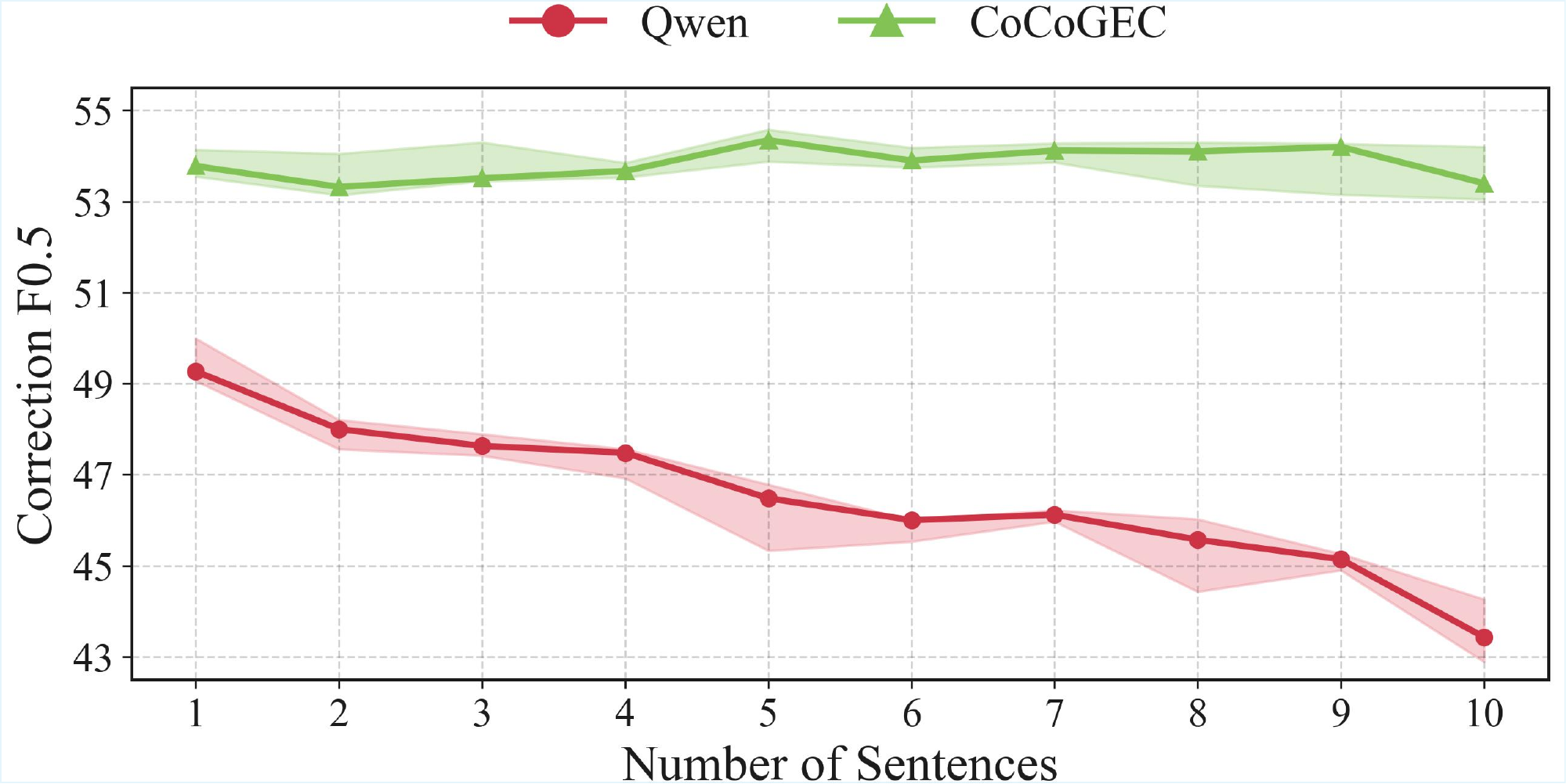}
\caption{Robustness to perturbation length on Qwen3-8B on TEM-8*. As context grows, the vanilla model degrades, whereas the \textsc{CoCoGEC}-trained model keeps higher and more stable $F_{0.5}$, indicating better long-range robustness.}

    \label{fig:context_length}
\end{figure}

\subsection{Robustness to Perturbation Length and Number}
\label{subsec:context_length}
To evaluate robustness under varying perturbation length and number, we use the LLM-based Qwen3-8B as the backbone.
We test on TEM-8 with sentence-level perturbations and increasing context length.
Specifically, for each error-containing sentence, we concatenate $k$ preceding and $k$ following sentences ($k \in \{1,\ldots,10\}$) as additional context, and we further apply word-level attacks at $m$ different positions ($m \in \{1,\ldots,5\}$) to vary the perturbation number.
We compare the vanilla Qwen3-8B model with Qwen3-8B trained with \textsc{CoCoGEC}.
Figure~\ref{fig:context_length} shows that the vanilla model degrades as the context window grows, whereas \textsc{CoCoGEC} maintains higher and more stable $F_{0.5}$, indicating reduced sensitivity to long-range context.
Table~\ref{tab:erroratk} reports results under word-level perturbations with increasing attack number: \textsc{CoCoGEC} consistently achieves higher $F_{0.5}$, TR, and SR across attack budgets, showing stronger tolerance to error-position shifts and denser error layouts.

\section{Conclusion}
\label{sec:con}
We examined the robustness gap in grammatical error correction and proposed \textsc{CoCoGEC}, a contextual counterfactual generation framework that preserves intended corrections while varying the surrounding discourse. Results on robustness-oriented and standard GEC benchmarks indicate that \textsc{CoCoGEC} improves correction accuracy and markedly enhances resilience to word-level and sentence-level contextual variations, pointing to contextual counterfactual generation as an effective data-centric approach to robust GEC.

\section*{Limitations}
\label{sec:limitations}
\textsc{CoCoGEC} currently relies on external large language models (LLMs) to generate span-controlled counterfactuals. This introduces a dependency on the particular LLM and prompting setup used for augmentation, and future work could explore lighter-weight or fully self-contained generators to further reduce this reliance.

\section*{Acknowledgement}
\label{sec:acknowledgement}
The authors would like to thank the anonymous reviewers for their insightful comments. This work is supported by the Chenguang Program of Shanghai Education Development Foundation and Shanghai Municipal Education Commission under Grant 24CGA26.

\bibliography{custom}

\appendix
\raggedbottom
\clearpage
\newpage
\section{Prompt Templates for Counterfactual Generation}
\label{app:prompts}


We instantiate our span-edit template over the training corpus to create a set of prompt--sentence pairs, denoted \texttt{gec\_examples\_with\_span}, for eliciting span-level edits from a large language model (LLM).
All prompts described in this section operate in a \emph{sentence} mode: given a full sentence, the LLM is asked either to fill a masked span or produce additional surrounding context.
This section summarizes the templates used in our counterfactual generation pipeline.

\subsection{Intra-sentential Editing}
\label{app:span-edit}

\paragraph{Global instruction.}
All span-edit prompts follow the ``Span-edit prompt (sentence mode)'' template below and differ only in the concrete \{sentence\} and span positions.

\paragraph{Unified span-edit template.}
We use a single template for replacement, deletion, and insertion.
The selected span (or insertion position) is masked with \texttt{[BLANK]}, and the LLM is required to output only the content that should fill \texttt{[BLANK]}.
An empty prediction corresponds to a deletion.
For insertion, we treat the insertion position as a zero-length span between two tokens and apply the same template.

\begin{tcolorbox}[
 colback=gray!5,
 colframe=gray!40!black,
 boxrule=0.5pt,
 sharp corners=south,
 title=Span-edit prompt (sentence mode),
 fontupper=\small\ttfamily\linespread{1.1}\selectfont
]
Task Instruction: Fill in the \texttt{[BLANK]} with a word or phrase, or leave it empty.

- Follow standard English grammar.

- No grammatical errors are allowed.

- Do not copy from the original sentence.

- The final sentence must be logically correct and sound natural to native speakers.

- Output \emph{only} the content that should replace \texttt{[BLANK]}. Do \emph{not} output the full sentence.

Sentence: \{sentence\}

\texttt{[BLANK]} should be:
\end{tcolorbox}

\subsection{Inter-sentential Expansion for Context Augmentation}
\label{app:expansion}

We perform inter-sentential expansion by generating a fluent context before and after an input sentence.
Given a training sample with source sentence $s$ and its correction $t$, we use an instruction-following LLM to generate a prefix $p$ and a suffix $q$ such that $p$ naturally leads into $s$ and $q$ naturally follows $s$, while keeping $s$ unchanged.
The expanded source and target are constructed as $p \oplus s \oplus q$ and $p \oplus t \oplus q$, respectively.
To improve diversity, we randomly sample the number of prefix/suffix sentences (e.g., 3--5) for each sample.
If the LLM output cannot be parsed into the required two-block format or violates constraints (e.g., repeating $s$), we fall back to the original sample without expansion.

\begin{tcolorbox}[
 colback=gray!5,
 colframe=gray!40!black,
 boxrule=0.5pt,
 sharp corners=south,
 title=Inter-sentential expansion prompt (prefix/suffix),
 fontupper=\small\ttfamily\linespread{1.1}\selectfont
]
You are given a sentence S in English. Your task is to write additional context BEFORE and AFTER S.

RULES:

1) In the \verb|<<<PREFIX>>>| block, write about \{n\_pre\} fluent English sentences that smoothly lead into S.

2) In the \verb|<<<SUFFIX>>>| block, write about \{n\_suf\} fluent English sentences that naturally follow S.

3) Do NOT change S at all. Keep it EXACTLY AS-IS.

4) Do NOT include S itself in the blocks.

5) Write only in English.

6) Output EXACTLY TWO blocks:

\verb|<<<PREFIX>>>|

...text before S...

\verb|<<<SUFFIX>>>|

...text after S...

S:
\{sentence\}
\end{tcolorbox}

\subsection{LLM-based Correction Prompt}
\label{app:gec-prompt}

For instruction-following LLMs used as correction models (or LLM-based scorers), we adopt the following sentence-level GEC prompt:

\begin{tcolorbox}[
 colback=gray!5,
 colframe=gray!40!black,
 boxrule=0.5pt,
 sharp corners=south,
 title=Grammar error correction prompt (GEC),
 fontupper=\small\ttfamily\linespread{1.1}\selectfont
]
You are an experienced English teacher who specializes in grammatical error correction (GEC). You are given exactly one sentence as input. Correct it with the fewest possible edits (minimal edit distance).

Requirements:

1) Correct grammar, spelling, and word choice only.
2) Keep the original structure and meaning; do not paraphrase or reorder.
3) If the sentence is already correct, return it unchanged.
4) Output format: return exactly one sentence on a single line, with no explanations or extra text.

Original: \{sentence\}

Corrected:
\end{tcolorbox}

\section{Additional Experimental Details}
\label{app:details}

\subsection{Dataset Statistics}
\label{app:stats}

Table~\ref{tab:split_stats_wide} reports corpus statistics for each split, including sentence counts, source-token volumes, and error-type breakdowns.

\begin{table}[htbp]
\centering
\setlength{\tabcolsep}{2pt}
\resizebox{\columnwidth}{!}{%
\begin{tabular}{lrrrrr}
\toprule
Properties & train & dev & test\_BEA-19 & test\_CoNLL-14 & test\_TEM-8 \\
\midrule
\#Sent & 22848 & 2538 & 3018 & 1578 & 1590 \\
\#Tokens(src) & 1.14M & 125.1K & 242.1K & 120.1K & 142.4K \\
Avg src length (tok) & 49.92 & 49.28 & 80.23 & 76.08 & 89.56 \\
Avg \#sentences in src & 2.87 & 2.85 & 4.90 & 4.51 & 4.78 \\
Errorful \% & 78.4\% & 78.4\% & 71.8\% & 87.5\% & 100.0\% \\
\#Edits per example & 1.999 & 1.963 & 2.095 & 2.369 & 2.029 \\
\#Missing errors per example & 0.373 & 0.357 & 0.217 & 0.278 & 0.282 \\
\#Redundant errors per example & 0.205 & 0.192 & 0.304 & 0.479 & 0.223 \\
\#Substitution errors per example & 1.407 & 1.394 & 1.560 & 1.606 & 1.524 \\
\#Word-Order errors per example & 0.014 & 0.019 & 0.014 & 0.007 & 0.000 \\
\bottomrule
\end{tabular}}
\caption{Dataset statistics across splits. Errorful \% is the percentage of examples with at least one ERRANT edit between (src, tgt). Sentence counts are reported using spaCy segmentation and a rule-based heuristic (punctuation/newline boundaries).}
\label{tab:split_stats_wide}
\end{table}
\subsection{Gold Validity Check}
\label{app:gold-validity}

After enforcing the edit-subset constraint ($E' \subseteq E$), we apply a lightweight gold-validity check to ensure that the revised target $t'$ remains valid for the perturbed source $c'$. We discard candidates if (i) $t'$ fails automatic grammaticality checking, (ii) a frozen external GEC verifier further edits $t'$, or (iii) $t'$ shows obvious semantic drift.

\paragraph{Quality-control pipeline.}
We further apply a filtering pipeline to ensure that generated contexts are valid distractors without introducing new errors or leaking the correction.

\paragraph{Grammaticality check (ERRANT).}
Using the same ERRANT-based edit extraction pipeline as in the main experiments, we verify that generated prefixes and suffixes introduce no additional grammatical errors. Most generated contexts pass this check, yielding a Context Robustness Score (CRS) of 99.6\%.

\paragraph{Fluency and coherence filtering.}
We use a perplexity filter to remove non-fluent generations and a semantic-similarity constraint to filter severe semantic drift, ensuring natural and coherent augmented contexts.

\paragraph{Leakage prevention.}
We remove candidates that may directly reveal the target correction, reducing reliance on accidental lexical cues.

\paragraph{Manual verification.}
Manual inspection of sampled cases shows that the retained counterfactuals generally preserve the original error pattern while varying only the surrounding context, supporting the validity of our pipeline.
\begin{table}[t]
\centering
\small
\begin{tabular}{l l}
\toprule
\textbf{Parameter} & \textbf{Value}\\
\midrule
Target modules   & all \\
LoRA rank     & 16 \\
Learning rate    & $5\times 10^{-5}$\\
Training epochs   & 2 \\
Per-device batch size  & 2 \\
Gradient accumulation steps & 8 \\
Warmup ratio    & 0.1 \\
Max sequence length  & 1024 \\
Scheduler     & cosine \\
Precision     & bf16 \\
\bottomrule
\end{tabular}
\caption{LoRA configuration for fine-tuning Qwen3-8B.}
\label{tab:lora}
\end{table}

\subsection{Hyperparameters}
\label{app:implementation}

For the LLM-based scorer, we fine-tune Qwen3-8B with LoRA~\citep{hu2021lora} using LLaMA-Factory~\citep{zheng2024llamafactory}; the LoRA configuration is summarized in Table~\ref{tab:lora}.

We also train and evaluate GECToR-large and T5-large backbones.
For GECToR-large, we follow the official implementation and training hyperparameters used in RobustGEC~\citep{zhang-etal-2023-robustgec}.
For T5-large, we follow the text-to-text training recipe of \citet{rothe-etal-2021-simple} and use the public \texttt{gec-t5} implementation.\footnote{\url{https://github.com/gotutiyan/gec-t5}}
For $\mathcal{DISCO}$ and TypeDA, we implement the methods as described in the original papers and run them with the authors' recommended hyperparameters and filtering rules, without additional tuning beyond adapting them to our GEC data.
Unless otherwise specified, all baseline settings are kept identical to the referenced implementations.
\begin{table}[htbp]
\centering
\setlength{\tabcolsep}{4pt}
\resizebox{\columnwidth}{!}{%
\begin{tabular}{l|ccc|ccc|ccc}
\toprule
\multirow{2}{*}{\textbf{Model}}
 & \multicolumn{3}{c|}{BEA-19 dev}
 & \multicolumn{3}{c|}{CoNLL-14 test}
 & \multicolumn{3}{c}{BEA-19 test} \\
 & P & R & $F_{0.5}$
 & P & R & $F_{0.5}$
 & P & R & $F_{0.5}$ \\
\midrule
GECToR
 & $68.8$ & $38.8$ & $59.6$
 & $75.4$ & $40.9$ & $64.5$
 & $79.0$ & $56.2$ & $73.1$ \\
+ \textsc{CoCoGEC}
 & $72.2$ & $42.1$ & $\mathbf{63.2}$
 & $72.6$ & $54.6$ & $\mathbf{68.2}$
 & $80.0$ & $57.2$ & $\mathbf{74.1}$ \\
\bottomrule
\end{tabular}%
}
\caption{Evaluating \textsc{GECToR}-large with and without \textsc{CoCoGEC} on its original BEA-19 and CoNLL-14 benchmarks.}
\label{tab:standard_gec}
\end{table}
\subsection{Standard GEC Performance on Original Benchmarks}
\label{app:standard-gec}
Table~\ref{tab:standard_gec} reports standard GEC results on the original BEA-19 dev/test and CoNLL-14 test sets.
Compared with the vanilla GECToR backbone, the \textsc{CoCoGEC}-augmented model attains similar or slightly higher $F_{0.5}$ scores on all three benchmarks (up to +3.7 on CoNLL-14), indicating that improving robustness on RobustGEC does not degrade conventional GEC performance.
\subsection{Supplementary Cross-Lingual Results on VisCGEC}
Following the reviewer suggestion on cross-lingual and cross-domain validation, we additionally evaluate COCOGEC on the Chinese VisCGEC benchmark. Results are shown in Table~\ref{tab:viscgec_appendix}. We follow the official VisCGEC evaluation protocol and compare the baseline and COCOGEC-augmented models under the same training and decoding settings, without additional model-specific tuning.

\begin{table}[h]
\centering
\small
\begin{tabular}{lccc}
\toprule
Method & Precision & Recall & $F_{0.5}$ \\
\midrule
Baseline & 32.30 & 20.23 & 28.86 \\
COCOGEC (ours) & \textbf{43.53} & 16.12 & \textbf{32.49} \\
\bottomrule
\end{tabular}
\caption{Supplementary results on the Chinese VisCGEC benchmark.}
\label{tab:viscgec_appendix}
\end{table}

As shown in Table~\ref{tab:viscgec_appendix}, COCOGEC also improves $F_{0.5}$ on VisCGEC, suggesting that the proposed context-decoupled counterfactual augmentation is not limited to the English RobustGEC setting and can generalize to a different language and benchmark.
\subsection{Context Robustness Metrics: CRS and P-CRS}
\label{app:crs-metric}

Following RobustGEC~\citep{zhang-etal-2023-robustgec}, we report two context-robustness metrics: Context Robustness Score (CRS) and Pair-wise Context Robustness Score (P-CRS).
Each GEC case contains one original sentence and a set of context-perturbed variants.
CRS measures strict stability: it counts a case as correct only if the model outputs exactly identical corrections for \emph{all} variants within the same case,
\begin{equation}
\mathrm{CRS}=\frac{\#\mathrm{Case}_C}{\#\mathrm{Case}_T}.
\end{equation}
P-CRS is more lenient and evaluates stability at the original$\Leftrightarrow$perturb pair level,
\begin{equation}
\mathrm{P\mbox{-}CRS}=\frac{\#\mathrm{P\mbox{-}sample}_C}{\#\mathrm{P\mbox{-}sample}_T}.
\end{equation}
For example, if a case has one original and five perturbed variants where four variants share the same correction as the original, then CRS is $0$ while P-CRS is $4/5$.

\begin{table}[H]
\centering
\small
\begin{tabular}{lcc}
\toprule
& \multicolumn{2}{c}{TEM-8} \\
\cmidrule(lr){2-3}
Method & CRS$\uparrow$ & P-CRS$\uparrow$ \\
\midrule
Seq2Edit (GECToR)  & $6.41$ & $64.30$ \\
\quad + CPR method  & $7.93$ & $72.22$ \\
\quad + TypeDA   & $3.39$ & $63.09$ \\
\quad + $\mathcal{DISCO}$   & $8.25$ & $69.77$ \\
\quad + \textsc{CoCoGEC}  & $\mathbf{12.45}$ & $\mathbf{72.52}$ \\
\midrule
Seq2Seq (T5)   & $0.75$ & $53.20$ \\
\quad + CPR method  & $2.26$ & $59.22$ \\
\quad + TypeDA   & $0.75$ & $53.13$ \\
\quad + $\mathcal{DISCO}$   & $0.75$ & $51.54$ \\
\quad + \textsc{CoCoGEC}  & $\mathbf{4.15}$ & $\mathbf{63.54}$ \\
\midrule
LLM(Qwen3-8B)     & $16.60$ & $67.62$ \\
\quad + CPR method  & $7.93$ & $72.22$ \\
\quad + TypeDA   & $3.01$ & $50.26$ \\
\quad + $\mathcal{DISCO}$   & $6.41$   & $55.24$ \\
\quad + \textsc{CoCoGEC}  & $\mathbf{16.98}$   & $\mathbf{72.52}$ \\
\midrule
GPT-4o    & $8.67$ & $\mathbf{69.57}$ \\
LLaMA3-8B     & $1.13$ & $56.83$ \\
\bottomrule
\end{tabular}
\caption{CRS and P-CRS on TEM-8. \textsc{CoCoGEC} improves robustness over vanilla GECToR and T5, while the effects on LLM-based baselines are mixed.}
\label{tab:crs}
\end{table}
\subsection{CRS and P-CRS on TEM-8}
\label{app:crs-results}

Table~\ref{tab:crs} summarizes CRS and P-CRS on TEM-8*.
For Seq2Edit (GECToR) and Seq2Seq (T5), 
\textsc{CoCoGEC} improves robustness over the corresponding vanilla models for both GECToR and T5. For GECToR, $\mathcal{DISCO}$ yields the highest CRS/P-CRS among the training variants, while \textsc{CoCoGEC} still provides a clear gain over the vanilla baseline. For Qwen3, \textsc{CoCoGEC} gives a modest improvement in both CRS and P-CRS, whereas other recipes exhibit trade-offs (e.g., CPR increases P-CRS but lowers CRS), suggesting that stability improvements for LLM-based correction may be sensitive to the training recipe.

\subsection{Case Study}
\label{app:case}

We present representative long-context cases to qualitatively compare model behaviors.
For each case, we report the original input, the expanded input, the gold correction, and model predictions from different systems (e.g., GECToR, T5, Qwen3, ChatGPT, and LLaMA).
\begin{table*}[!t]
\centering
\scriptsize
\setlength{\tabcolsep}{2pt}
\renewcommand{\arraystretch}{1.02}
\begin{tabularx}{\textwidth}{@{}>{\raggedright\arraybackslash}p{0.17\textwidth}>{\raggedright\arraybackslash}X@{}}
\multicolumn{2}{@{}l@{}}{\textbf{Case Study: Contextual counterfactuals for ``an sad person''}}\\[0.4ex]
\hline
\textbf{Line} & \textbf{Utterance} \\
\hline

\textbf{Original $s$} &
\emph{The key to \textbf{\textcolor{red}{comfort}} a person is to try and avoid a debate over \textbf{\textcolor{red}{if}} your loved one is sick and instead \textbf{\textcolor{red}{of look}} for common ground.} \\[0.25ex]

\textbf{Original $t$} &
\emph{The key to \textbf{\textcolor{mygreen}{comforting}} a person is to try and avoid a debate over \textbf{\textcolor{mygreen}{whether}} your loved one is sick and instead \textbf{\textcolor{mygreen}{look}} for common ground.} \\[0.25ex]

\textbf{Counterfactual $s'$} &
\emph{The key to \textbf{\textcolor{red}{comfort}} a \underline{sad} person is to try and avoid a debate over \textbf{\textcolor{red}{if}} your loved one is sick and instead \textbf{\textcolor{red}{of look}} for common ground.} \\[0.25ex]

\textbf{Counterfactual $t'$} &
\emph{The key to \textbf{\textcolor{mygreen}{comforting}} a \underline{sad} person is to try and avoid a debate over \textbf{\textcolor{mygreen}{whether}} your loved one is sick and instead \textbf{\textcolor{mygreen}{look}} for common ground.} \\[0.25ex]

\textbf{Difference $s$ vs.\ $s'$} &
\emph{The counterfactual $s'$ inserts the descriptor \underline{``sad''} before ``person'', while the local error pattern around \textbf{\textcolor{red}{comfort}} / \textbf{\textcolor{red}{if}} / ``instead \textbf{\textcolor{red}{of look}}'' remains unchanged.} \\
\hline

\textbf{GECToR-large on $s$} &
\emph{The key to \textbf{\textcolor{mygreen}{comforting}} a person is to try and avoid a debate over \textbf{\textcolor{red}{if}} your loved one is sick and instead \textbf{\textcolor{red}{of looking}} for common ground.} \\[0.25ex]

\textbf{GECToR-large on $s'$} &
\emph{The key to \textbf{\textcolor{mygreen}{comforting}} a sad person is to try and avoid a debate over \textbf{\textcolor{red}{if}} your loved one is sick and instead \textbf{\textcolor{red}{of looking}} for common ground.} \\[0.25ex]

\textbf{T5-large on $s$} &
\emph{The key to \textbf{\textcolor{mygreen}{comforting}} a person is to try and avoid a debate over \textbf{\textcolor{red}{if}} your loved one is sick and instead \textbf{\textcolor{red}{of looking}} for common ground.} \\[0.25ex]

\textbf{T5-large on $s'$} &
\emph{The key to \textbf{\textcolor{mygreen}{comforting}} a sad person is to try and avoid a debate over \textbf{\textcolor{red}{if}} your loved one is sick and instead \textbf{\textcolor{red}{of looking}} for common ground.} \\[0.25ex]

\textbf{Qwen3-8B on $s$} &
\emph{The key to \textbf{\textcolor{mygreen}{comforting}} a person is to try and avoid a debate over \textbf{\textcolor{mygreen}{whether}} your loved one is sick and instead \textbf{\textcolor{red}{of looking}} for common ground.} \\[0.25ex]

\textbf{Qwen3-8B on $s'$} &
\emph{The key to \textbf{\textcolor{mygreen}{comforting}} a sad person is to try and avoid a debate over \textbf{\textcolor{mygreen}{whether}} your loved one is sick and instead \textbf{\textcolor{red}{of looking}} for common ground.} \\[0.25ex]

\textbf{GPT-4o on $s$} &
\emph{The key to \textbf{\textcolor{mygreen}{comforting}} a person is to try \textbf{\textcolor{red}{to}} avoid a debate over \textbf{\textcolor{mygreen}{whether}} your loved one is sick and instead \textbf{\textcolor{mygreen}{look}} for common ground.} \\[0.25ex]

\textbf{GPT-4o on $s'$} &
\emph{The key to \textbf{\textcolor{mygreen}{comforting}} a sad person is to try \textbf{\textcolor{red}{to}} avoid a debate over \textbf{\textcolor{mygreen}{whether}} your loved one is sick and instead \textbf{\textcolor{red}{of looking}} for common ground.} \\[0.25ex]

\textbf{LLaMA3-8B on $s$} &
\emph{The key to \textbf{\textcolor{mygreen}{comforting}} a person is to try \textbf{\textcolor{red}{to}} avoid \textbf{\textcolor{red}{debating}} \textbf{\textcolor{mygreen}{whether}} your loved one is sick and instead \textbf{\textcolor{mygreen}{look}} for common ground.} \\[0.25ex]

\textbf{LLaMA3-8B on $s'$} &
\emph{The key to \textbf{\textcolor{mygreen}{comforting}} a sad person is to try \textbf{\textcolor{red}{to}} avoid \textbf{\textcolor{red}{debating}} \textbf{\textcolor{red}{if}} your loved one is sick and instead \textbf{\textcolor{mygreen}{look}} for common ground.} \\[0.25ex]

\textbf{Qwen3-8B+COCOGEC on $s$} &
\emph{The key to \textbf{\textcolor{mygreen}{comforting}} a sad person is to try and avoid a debate over \textbf{\textcolor{mygreen}{whether}} your loved one is sick and instead \textbf{\textcolor{mygreen}{look}} for common ground.} \\[0.25ex]

\textbf{Qwen3-8B+COCOGEC on $s'$} &
\emph{The key to \textbf{\textcolor{mygreen}{comforting}} a person is to try and avoid a debate over \textbf{\textcolor{mygreen}{whether}} your loved one is sick and instead \textbf{\textcolor{mygreen}{look}} for common ground.} \\
\hline
\end{tabularx}
\caption{Line-by-line case study of a contextual counterfactual pair $(s,s')$ for ``an sad person''. Red bold spans mark residual errors relative to the gold targets, while dark-green bold spans highlight canonical corrections.}
\label{tab:case-study-depression}
\end{table*}

\end{document}